\documentclass[letterpaper,table]{article} %
\usepackage{aaai24}  %
\usepackage{times}  %
\usepackage{helvet}  %
\usepackage{courier}  %
\usepackage[hyphens]{url}  %
\usepackage{graphicx} %
\usepackage{natbib}  %
\usepackage{caption} %
\usepackage{newfloat}
\usepackage{listings}

\nocopyright

\usepackage[switch, modulo]{lineno}

\usepackage{amsmath}
\usepackage{amsthm}
\usepackage{amssymb}
\usepackage{booktabs}
\usepackage{stmaryrd}
\usepackage{algorithm}
\usepackage{algpseudocode}
\usepackage[inline]{enumitem}
\usepackage{nicefrac}
\usepackage{todonotes}
\usepackage{xcolor}

\usepackage{tikz}

\title{Consolidating LAMA with Best-First Width Search}
\author{
  Augusto B. Corr\^{e}a\textsuperscript{\rm 1},
  Jendrik Seipp\textsuperscript{\rm 2}
}
\affiliations {
  \textsuperscript{\rm 1}University of Basel, Switzerland \\
  \textsuperscript{\rm 2}Link\"{o}ping University, Sweden\\
  augusto.blaascorrea@unibas.ch, jendrik.seipp@liu.se
}

\begin{document}

\maketitle

\newtheorem{definition}{Definition}
\newtheorem{theorem}{Theorem}
\newtheorem{corollary}{Corollary}
\newcommand{\tup}[1]{\ensuremath{\langle #1 \rangle}}

\newcommand{\egcite}[1]{\citep[e.g.,][]{#1}}
\newcommand{\cfcite}[1]{\citep[c.f.,][]{#1}}
\newcommand{\inlinecite}[1]{\citet{#1}}

\newcommand{\sasplus}{\ensuremath{\textup{SAS}^+{}}}
\newcommand{\variables}{\ensuremath{\mathcal{V}}}
\newcommand{\dom}{\ensuremath{\textit{dom}}}
\newcommand{\vars}{\ensuremath{\textit{vars}}}
\newcommand{\actions}{\ensuremath{\mathcal{A}}}
\newcommand{\statespace}{\ensuremath{\mathcal{S}}}
\newcommand{\states}{\ensuremath{S}}
\newcommand{\pre}{\ensuremath{\textit{pre}}}
\newcommand{\eff}{\ensuremath{\textit{eff}}}
\newcommand{\cost}{\ensuremath{\textit{cost}}}
\newcommand{\init}{\ensuremath{s_I}}
\newcommand{\goal}{\ensuremath{G}}
\newcommand{\goals}{\ensuremath{S_G}}
\newcommand{\app}[2]{\ensuremath{#1\llbracket #2 \rrbracket}}
\newcommand{\suc}{\ensuremath{\textit{succ}}}
\newcommand{\facts}{\ensuremath{F}}

\newcommand{\hadd}{\ensuremath{h^\text{add}}}
\newcommand{\hff}{\ensuremath{h^\text{FF}}}
\newcommand{\hlm}{\ensuremath{h^\text{LM}}}

\newcommand{\cmark}{\ensuremath{\checkmark}}%

\newcommand{\augusto}[1]{\todo[inline,color=teal!50,caption={}]{Augusto: #1}}
\newcommand{\jendrik}[1]{\todo[inline]{Jendrik: #1}}
\newcommand{\acronym}[1]{\underline{#1}}
\renewcommand{\acronym}[1]{#1}

\begin{abstract}

One key decision for heuristic search algorithms is how to balance exploration
and exploitation. In classical planning, novelty search has come out as the most
successful approach in this respect. The idea is to favor states that contain
previously unseen facts when searching for a plan. This is done by maintaining a
record of the tuples of facts observed in previous states. Then the novelty of a
state is the size of the smallest previously unseen tuple. The most successful
version of novelty search is best-first width search (BFWS), which combines
novelty measures with heuristic estimates. An orthogonal approach to balance
exploration-exploitation is to use several open-lists. These open-lists are
ordered using different heuristic estimates, which diversify the information
used in the search. The search algorithm then alternates between these
open-lists, trying to exploit these different estimates. This is the approach
used by LAMA, a classical planner that, a decade after its release, is still
considered state-of-the-art in agile planning. In this paper, we study how to
combine LAMA and BFWS. We show that simply adding the strongest open-list used
in BFWS to LAMA harms performance. However, we show that combining only parts of
each planner leads to a new state-of-the-art agile planner.

\end{abstract}

\section{Introduction}

\emph{Agile planning} involves solving planning tasks as fast as possible, with
little or no consideration for plan quality. While this might not be suitable
for some domains, it is still an important question in general, and it is
closely related to the problem of deciding plan existence
\egcite{bylander-aij1994}. Since 2014, there have been dedicated tracks for
agile planning in the classical part of the International Planning Competition
(IPC). Usually, planners are given a $5$ minutes per task, and they are scored
based on how quickly they solve the task.

In the recent IPC 2023, LAMA \cite{richter-et-al-ipc2011}, a 15 year-old planner
that was included as a baseline, obtained a higher total agile score than all
competitors, including the winner DecStar
\cite{gnad-et-al-ipc2023b}.\footnote{When we mention LAMA in our paper, we refer
to the \emph{first iteration} of LAMA, which is responsible for all its
coverage. Further iterations are used to improve plan quality.} A similar
situation occurred in the IPC 2018, where the LAMA baseline scored second place.

LAMA uses an \emph{alternation of open-lists} \cite{roeger-helmert-icaps2010}
together with \emph{deferred evaluation} and \emph{preferred operators}
\cite{richter-helmert-icaps2009} to find a plan quickly. This alternation allows
LAMA to balance between multiple heuristic estimates and allowed it to win the
Satisficing Track of the IPC in 2008. Even though the LAMA code base is
continuously improved, the core behavior remains unchanged. This raises the
question of how we can advance the state of the art and obtain a planner that
finds plans faster than LAMA.

In the IPC 2018 Agile Track, the only planner to outperform LAMA was based on
best-first width search (BFWS)
\cite{lipovetzky-geffner-aaai2017,frances-et-al-ipc2018}. The idea of BFWS is to
expand states that are \emph{novel} \cite{lipovetzky-geffner-ecai2012}. There
are multiple ways to define the novelty of a state
\egcite{katz-et-al-icaps2017}, but the key idea is that the state contains a
tuple of atoms which has not been encountered before by the search. When
combined with heuristic estimates, BFWS achieves a good exploration-exploitation
trade-off which helps find plans quickly.

In this paper, we aim to combine the advantages of LAMA and BFWS. Our basic idea
is very simple: we add the open-list used by BFWS (more precisely, BFWS($f_6$))
to LAMA, as one of its alternated open-lists. One could expect that this
combination would result in a faster planner because classical planning has a
history of ``\emph{the more, the merrier}'': more open-lists
\egcite{roeger-helmert-icaps2010,correa-et-al-ipc2023c}, more heuristics
\egcite{franco-et-al-ijcai2017,seipp-et-al-jair2020}, more tie-breakers
\egcite{asai-fukunaga-jair2017}, more planners
\egcite{helmert-et-al-icaps2011wspal,cenamor-et-al-jair2016}, all seem to
improve coverage of the planners.

It turns out though, that this combination actually yields a \emph{worse} planner than LAMA and
BFWS($f_6$) alone. However, a detailed ablation study reveals that
\emph{removing features} from LAMA and BFWS drastically speeds up the planner.
In fact, by removing entire open-lists of the LAMA configuration, and discarding
function estimates and tie-breakers from BFWS($f_6$), improves the coverage and
IPC agile score compared to the original systems. The resulting planner also
outperforms all agile competitors of the last IPCs by a large margin, with an
IPC agile score $13\%$ higher than the second best method. Overall, our new
method establishes a new state of the art for agile planning.

\section*{Background}

A \emph{state space} is a tuple $\statespace = \tup{\states, \init , \goal,
\suc}$, where $\states$ is a set of \emph{states}, $\init \in \states$ is the
\emph{initial state}, $\goal$ is the \emph{goal description}, $\suc$ is a
\emph{successor function} mapping each state to a finite (possible empty) set of
successor states. A state is a set of \emph{facts}. If a state $s$ contains fact
$f$ or a set of facts $\facts$, we say that $s$ \emph{satisfies} $f$ or
$\facts$. It is sufficient to consider $S$ as the minimal set where $\init \in
S$, and $\suc(s) \subseteq S$ for each $s \in S$. The goal $G$ is also a set of
facts. A state $s_*$ is a \emph{goal state} if $G \subseteq s_*$.

A sequence of states $\tau = \tup{s_0, \ldots , s_n}$ is a \emph{path} from
$s_0$ to $s_n$ in $S$ if $s_i \in \suc(s_{i-1})$ for $i \in \{1, \ldots , n\}$.
Path $\tau$ is an \emph{$s$-plan} if $s_0 = s$ and $\goal \subseteq s_n$ and a
\emph{plan} for $\statespace$ if $s_0 = \init$.

We consider \emph{agile planning}, the problem of computing plans as
fast as possible, without caring about their \emph{quality} (i.e., number of
states visited by the plan).

A common method to find plans is via \emph{heuristic search}. A \emph{heuristic}
is a function $h: S \to \mathbb{R}^+_0 \cup \{\infty\}$. It estimates the length
of an $s$-plan for states $s \in \states$. Heuristic search algorithms start
from $\init$ and explore states guided by some heuristic $h$, preferring states
$s$ with low $h(s)$ values. Examples of strong heuristics for agile planning are
\hadd~\cite{bonet-geffner-aij2001}, \hff~\cite{hoffmann-nebel-jair2001}, and
\hlm \cite{richter-et-al-aaai2008}. We assume familiarity with common search
algorithms such as greedy best-first search \cite{doran-michie-rsl1966}.

Instead of being guided by a heuristic, BFS($w$)
selects states for expansion based on
their \emph{novelty}, preferring states with low $w$ values
\cite{lipovetzky-geffner-ecai2012,lipovetzky-geffner-aaai2017}. The novelty
$w(s)$ of a state $s$ is the size of the smallest set of facts $\facts$ such
that $s$ is the first state visited that satisfies $\facts$.

This simple scheme can be turned into state-of-the-art \emph{best-first width
search} (BFWS) algorithms by extending it with
\emph{partition functions}
\cite{lipovetzky-geffner-aaai2017,frances-et-al-ijcai2017,frances-et-al-ipc2018}.
For BFWS, the novelty $w_{\tup{h_1, \ldots, h_n}}(s)$ of a state $s$ given the
\emph{partition functions} $\tup{h_1, \ldots, h_n}$ is the size of the smallest
set of facts $\facts$ such that $s$ is the first evaluated state that subsumes
$\facts$, among all states $s'$ visited before $s$ for which $h_i(s) = h_i(s')$
for $1 \leq i \leq n$. In practice, these planners only evaluate novelty up to a
bound $k$, where usually $k=2$. If a state $s$ has no novel tuple of size $k$ or
less, then $w(s) = k+1$.

\section*{Balancing Exploration and Exploitation}

One important design choice of a planner is how it balances \emph{exploration}
and \emph{exploitation}. Exploration techniques search parts of the state space
that have not yet been visited. Exploitation techniques prefer going into parts
that are considered more promising by some metric. For example, choosing the
next expanded state at random is a form of exploration; choosing it based on a
heuristic is exploitation.

Modern planners usually mix both of them. A common technique is to keep several
open-lists during search, each one guided by a different heuristic
\cite{roeger-helmert-icaps2010}. Some lists are even \emph{incomplete} since
they only retain a subset of the generated states, e.g., states generated via
\emph{preferred operators}
\cite{hoffmann-nebel-jair2001,richter-helmert-icaps2009}. The simplest yet most
successful method for combining multiple open-lists is \emph{alternation}
\cite{helmert-icaps2004,helmert-jair2006}. An alternation-based search algorithm
maintains $n$ open-lists, where the $i$-th open-list is ordered by some
heuristic $h_i$. We denote it as $[h_1, \ldots, h_n]$. The search alternates
between the open-lists in a round-robin fashion: first it expands the best state
according to $h_1$, adds all successors to all (or some of the) open-lists, then
it expands the best state according to $h_2$, and so on. In iteration $n+1$ it
expands from $h_1$ again.

Alternation is one of the main building blocks used by the LAMA planner
\cite{richter-westphal-ipc2008,richter-westphal-jair2010,richter-et-al-ipc2011}.
LAMA keeps four open-lists: $[\hff, \hff_+, \hlm, \hlm_+]$, where $h_+$ denotes
an open-list ordered by $h$ but only containing states reached via preferred
operators. LAMA is based on the Fast Downward planning system
\cite{helmert-jair2006} and won the IPC editions of 2008 and 2011. Until today,
it is considered state-of-the-art in agile planning.

An orthogonal way of combining exploration and exploitation is by using
\emph{tiebreakers} \cite{roeger-helmert-icaps2010}. A tiebreaking open-list
$\tup{h_1, \ldots, h_n}$ uses a ranking over $n$ heuristics. It selects states
based on $h_i$ and, if there is a tie, breaks this tie using $h_{i+1}$. It keeps
only a single open-list, but the order of this list is defined by multiple
heuristics. Throughout the paper, we assume that if all $h_1, \ldots, h_n$ are
tied, then remaining ties are broken by $g$-value. If ties persist, then we
assume a FIFO ordering.

While LAMA opted for an alternation open-list, the more sophisticated versions
of BFWS use tiebreaking \cite{lipovetzky-geffner-aaai2017}. In general,
BFWS($f$) orders its open-list by $f = \tup{f_1, \ldots, f_n}$. The strongest
version of BFWS is BFWS($f_6$), where the open-list is ordered by $f_6 =
\tup{w_{\tup{\hlm,\hff}}, \mathit{pref}, \hlm, w_{\tup{\hff}}, \hff}$ where
$\mathit{pref}$ is an indicator function yielding 1 for states reached via a
preferred operator. Simpler versions include BFWS($f_4$), where $f_4 =
\tup{w_{\tup{\hlm,\hff}}, \hlm, \hff}$, and BFWS($f_2$), where $f_2 =
\tup{w_{\tup{\hff}}, \hff}$.

A BFWS-based planner, BFWS-Preference, won the Agile Track of the IPC 2018
\cite{frances-et-al-ipc2018}. This planner is the only one to achieve a higher
agile score than LAMA in any of the IPCs, which is remarkable given that LAMA
only served as a baseline planner.

\section*{Combining LAMA and BFWS}

LAMA and BFWS present distinct ways of combining exploration and
exploitation. But they do have more similarities than initially meets the eye.
If we consider BFWS($f_6$), then both planners use exactly the same information
(with the exception of the novelty measures): \hff, \hlm, and preferred
operators. The difference is in the way the planners process the information and
the question is how we can combine the advantages of both in a single
planner. Arguably the most direct approach to combine both planners is to use
the tiebreaking open-list of BFWS($f_6$) as an additional open list in LAMA. We
call this modification LAMA-W($f_6$).

\begin{table*}[t!]
  \setlength{\tabcolsep}{7pt}
\centering
\begin{tabular}{@{}lrrrrrrrrr@{}}
                         & \makebox[0pt][r]{\rotatebox{-22}{LAMA}} & \makebox[0pt][r]{\rotatebox{-22}{BFWS($f_6$)}} & \makebox[0pt][r]{\rotatebox{-22}{LAMA-W($f_6$)}} & \makebox[0pt][r]{\rotatebox{-22}{LAMA-W($f_4$)}} & \makebox[0pt][r]{\rotatebox{-22}{LAMA-W($f_2^{\hff}$)}} & \makebox[0pt][r]{\rotatebox{-22}{LAMA-W($f_2^{\hlm}$)}} & \makebox[0pt][r]{\rotatebox{-22}{LAMA-W($w_{\tup{\hff}}$)}} & \makebox[0pt][r]{\rotatebox{-22}{LAMA-W($w_{\tup{\hlm}}$)}} & \makebox[0pt][r]{\rotatebox{-22}{LAMA-W($w_{\tup{}}$)}} \\ \midrule
{Coverage}         & 2081                                    & 2042                                           & 2029                                             & 2037                                             & 2047                                                    & \textbf{2113}                                           & 1957                                                  & 2096                                                  & 2094                                                     \\
{Expansions Score}       & 1629.31                                 & \textbf{1687.50}                               & 1359.53                                          & 1363.68                                          & 1345.63                                                 & 1643.34                                                 & 1238.47                                               & 1630.12                                               & 1634.12                                                  \\
{Agile Score}            & 1737.43                                 & 1581.45                                        & 1593.69                                          & 1600.41                                          & 1587.31                                                 & \textbf{1751.11}                                        & 1509.52                                               & 1740.41                                               & 1739.66                                                  \\
\bottomrule
\end{tabular}
\caption{Scores for the baselines, LAMA, BFWS($f_6$) and LAMA-W($f_6$),
  and for simplifications of LAMA-W($f_6$).}
\label{tab:ablation-A}
\end{table*}

Before we evaluate this new planner, we present the experimental setup used
throughout the paper. We implemented LAMA-W($f_6$) and BFWS($f_6$), alongside
the simpler BFWS variants, within the Scorpion planning system
\cite{seipp-et-al-jair2020}, which is an extension of Fast Downward
\cite{helmert-jair2006}. All our alternation-based methods use the default
boosting parameters of LAMA \cite{richter-westphal-jair2010}: whenever the search
encounters a state $s$ with a new lowest heuristic value $h(s)$ for any heuristic $h$,
all preferred operator queues receive a \emph{boost} value of $1000$.
This incurs that for the next $1000 \cdot n$ expansions only states from
preferred operator queues are considered, where $n$ is the number of such queues.

For running our experiments, we use Downward Lab \cite{seipp-et-al-zenodo2017}
on AMD EPYC 7742 processors running at 2.25~GHz.  We use the same limits as the
IPC 2023 Agile Track: a 5 minutes time limit, and 8~GiB of memory per task. Our
benchmark set consists of 2426 tasks from all IPCs (1998--2023). We omit domains
containing derived predicates, as some of the planners do not support them. We
also use the $h^2$-preprocessor for all planner runs
\cite{alcazar-torralba-icaps2015}.
We employ three main scores to evaluate planner performance: coverage
score,\footnote{The coverage score is 1 for solved tasks and 0 otherwise.}
expansions score, and agile score.\footnote{The expansions score is based on
the number of expanded states, while the agile score is based on the runtime
\cite{richter-helmert-icaps2009}. Performance better than a lower bound (100
states for expansions and 1 second for runtime) counts as $1$. Performance worse
than an upper bound $U$ ($10^6$ states for expansions and $300$s for runtime)
counts as $0$. We interpolate intermediate values with a
logarithmic function: $1 - \nicefrac{\log(x)}{\log(U)}$ where $x$ is the number
of expansions/runtime in seconds.} The total score is the sum over all tasks.

The first three columns of Table~\ref{tab:ablation-A} compare LAMA and
BFWS($f_6$) with LAMA-W($f_6$). We see that the new algorithm solves fewer tasks
and needs more expansions than both LAMA and BFWS($f_6$) in almost all metrics,
but its agile score is slightly higher than the one by BFWS($f_6$). The low
expansions score indicates that the addition of the new open-list makes the
search less informed. This is somewhat counter-intuitive because in many cases
combining different evaluators within the same search algorithm decreases the
number of expansions \cfcite{correa-et-al-ipc2023c}.

To understand what exactly has such a negative impact on performance, we
decomposed the features of LAMA-W($f_6$). Our first step is to iteratively simplify
the new open list by removing its subsets of its features (partition functions,
tie-breakers). We define the following variations:
\begin{enumerate}
\item LAMA-W($f_4$), where $f_4
= \tup{w_{\tup{\hlm, \hff}}, \hlm, \hff}$;
\item LAMA-W($f_2^h$). where $f_2^h = \tup{w_{\tup{h}}, h}$. We tested two different versions: $h =
  \hff$ and $h = \hlm$;
\item LAMA-W($w_{\tup{h}}$) uses the open-list of BFWS($w_{\tup{h}}$)
  \cite{lipovetzky-geffner-aaai2017}, a best-first width search with a single
  open-list ordered by $w_{\tup{h}}$ and breaking ties by accumulated cost
  $g$. When $h = \tup{}$ we do not use any partition function.
\end{enumerate}

Table~\ref{tab:ablation-A} shows the results for all methods. From left to
right, the novelty-based open list becomes increasingly simple. There are two
immediate observations to be made: first, simpler approaches, such as
LAMA-W($f_2^{\hlm}$) and LAMA-W($w_{\tup{}}$), have much better coverage and
agile score than LAMA-W($f_6$); second, using $\hlm$ in the novelty open-list is
consistently superior to $\hff$. LAMA-W($f_2^{\hlm}$) improves over LAMA and
LAMA-W($f_6$) in all measured criteria. It is also superior to BFWS($f_6$) in
coverage and agile scores.

\begin{figure}
\centering
\includegraphics[scale=.625]{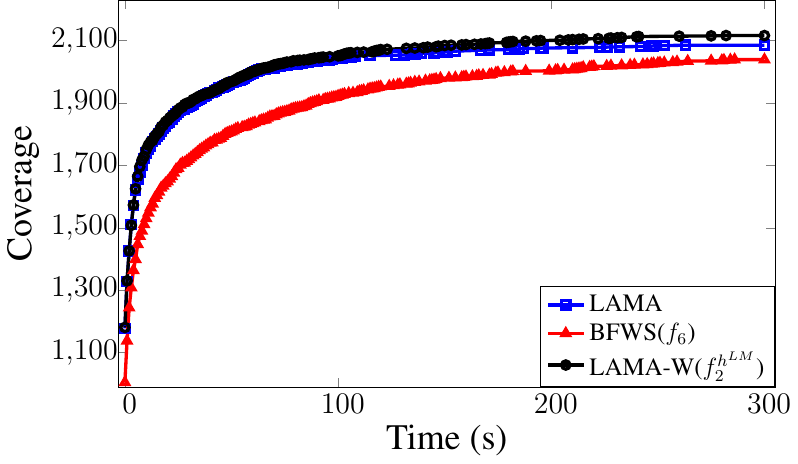}
\caption{Coverage over time for LAMA, BFWS($f_6$), and LAMA-W($f_2^{\hlm}$).}
\label{fig:cactus-baseline}
\end{figure}

Figure~\ref{fig:cactus-baseline} shows the coverage over time for LAMA,
BFWS($f_6$), and LAMA-W($f_2^{\hlm}$). Albeit having a better expansions score,
BFWS($f_6$) is slower than the other two methods. LAMA and LAMA-W($f_2^{\hlm}$)
run similarly fast, and the difference in the curves is only perceptible when
very close to the time limit. This indicates that LAMA-W($f_2^{\hlm}$) has an
edge over LAMA in larger tasks, where the extra exploration added by the novelty
open-list helps.

\newcommand{\checked}{\multicolumn{1}{c}{\cmark}}
\begin{table*}[t!]
  \setlength{\tabcolsep}{2pt}
\small
\begin{center}
\begin{tabular}{@{}lrrrrrrrrrrrrrrrr@{}}
  &  &        &        &        &        &        &        &        & & & & & & & & \multicolumn{1}{l}{\scriptsize LAMA-} \\
  &  &        &        &        &        &        &        &        & & & & & & \multicolumn{1}{l}{\scriptsize NOLAN} & & {\scriptsize W($f_2^{\hlm}$)} \\
\toprule
$\hff$   &  &        &        &        &        &        &        &        & \checked & \checked & \checked & \checked & \checked & \checked & \checked & \checked \\
$\hff_+$ &  &        &        &        & \checked & \checked & \checked & \checked &        &        &        &        & \checked & \checked & \checked & \checked \\
$\hlm$ &  &        & \checked & \checked &        &        & \checked & \checked &        &        & \checked & \checked &        &        & \checked & \checked \\
$\hlm_+$   &  & \checked &        & \checked &        & \checked &        & \checked &        & \checked &        & \checked &        & \checked &        & \checked \\
\midrule
{Coverage}               & 1668                                             & 2001                                                     & 1803                                                   & 2002                                                           & 2081                                                     & 2112                                                             & 2081                                                           & \textbf{2123}                                                          & 2023                                                   & 2056                                                           & 2013                                                         & 2052                                                                 & 2076                                                           & 2120                                                                   & 2074                                                                 & 2113                                                                         \\
{Exp. Score}             & 1029.7                                          & 1449.3                                                  & 1113.0                                                & 1444.2                                                        & 1533.4                                                  & 1631.6                                                          & 1536.7                                                        & 1642.5                                                                & 1332.9                                                & 1502.6                                                        & 1333.8                                                      & 1499.5                                                              & 1539.6                                                        & \textbf{1644.0}                                                                & 1535.8                                                              & 1643.3                                                             \\
{Agile Score}                  & 1278.6                                          & 1632.3                                                  & 1380.7                                                & 1627.8                                                        & 1694.8                                                  & 1744.3                                                          & 1699.9                                                        & 1753.9                                                                & 1581.1                                                & 1687.2                                                        & 1577.7                                                      & 1681.2                                                              & 1699.5                                                        & \textbf{1756.3}                                                       & 1696.5                                                              & 1751.1                                                                      \\
\bottomrule
 \end{tabular}
\caption{Results for the consolidation of LAMA open-lists in
  LAMA-W($f_2^{\hlm}$) (last column). The first four rows correspond to the four original open-lists in LAMA and the active open-lists are marked with $\cmark$. All configurations include the $f_2^{\hlm}$ open-list.}
\label{tab:ablation-B}
\end{center}
\end{table*}

\begin{table}[t!]
\begin{center}
  \begin{tabular}{@{}lrrr}
    \toprule
                                & LAMA & NOLAN         & Diff.                       \\ \midrule
barman-sat14        (20)        & 19   & 20            & \cellcolor{green!25} +1     \\
caldera-split-sat18    (20)     & 6    & 10            & \cellcolor{green!25}+4      \\
cavediving-14          (20)     & 7    & 8             & \cellcolor{green!25}+1      \\
data-network-sat18  (20)        & 11   & 15            & \cellcolor{green!25}+4      \\
depot                      (22) & 20   & 22            & \cellcolor{green!25}+2      \\
freecell                   (80) & 79   & 80            & \cellcolor{green!25}+1      \\
logistics98                (35) & 34   & 35            & \cellcolor{green!25}+1      \\
maintenance-sat14      (20)     & 11   & 12            & \cellcolor{green!25}+1      \\
nomystery-sat11     (20)        & 13   & 17            & \cellcolor{green!25}+4      \\
nurikabe-sat18         (20)     & 16   & 17            & \cellcolor{green!25}+1      \\
pathways                   (30) & 23   & 25            & \cellcolor{green!25}+2      \\
pipesworld-notankage       (50) & 44   & 45            & \cellcolor{green!25}+1      \\
pipesworld-tankage         (50) & 41   & 42            & \cellcolor{green!25}+1      \\
ricochet-robots-sat23  (20)     & 6    & 14            & \cellcolor{green!25}+8      \\
rubiks-cube-sat23      (20)     & 20   & 17            & \cellcolor{red!15}-3        \\
snake-sat18         (20)        & 4    & 9             & \cellcolor{green!25}+5      \\
storage                    (30) & 19   & 26            & \cellcolor{green!25}+7      \\
tetris-sat14        (20)        & 17   & 20            & \cellcolor{green!25}+3      \\
thoughtful-sat14    (20)        & 16   & 18            & \cellcolor{green!25}+2      \\
tidybot-sat11       (20)        & 18   & 20            & \cellcolor{green!25}+2      \\
transport-sat11     (20)        & 15   & 18            & \cellcolor{green!25}+3      \\
transport-sat14     (20)        & 11   & 12            & \cellcolor{green!25}+1      \\
trucks                   (30)   & 15   & 17            & \cellcolor{green!25}+2      \\
trucks-strips              (30) & 14   & 16            & \cellcolor{green!25}+2      \\
visitall-sat11      (20)        & 20   & 19            & \cellcolor{red!15}-1        \\
visitall-sat14      (20)        & 20   & 4             & \cellcolor{red!15}-16       \\
\midrule
Other 55 domains   (1729)       & 1562 & 1562          & 0                           \\
\midrule
Total (2426)                    & 2081 & \textbf{2120} & \cellcolor{green!25}\bf +39 \\
    \bottomrule
    \end{tabular}
\caption{Per domain coverage for LAMA and NOLAN. We only show domains where
  coverage differs. The last column shows the coverage gain/loss for NOLAN.}
\label{tab:domains}
\end{center}
\end{table}

\begin{table}[t!]
\begin{center}
  \begin{tabular}{@{}lrr@{}}
    \toprule
           & Coverage &  Agile Score \\ \midrule
NOLAN      & \bf 2120 &  \bf 1756.3  \\
GBFS-RSL   & 2049     &  1557.3      \\
Dual-BFWS  & 1986     &  1537.4      \\
OLCFF      & 2024     &  1504.9      \\
BFWS-Pref. & 1927     &  1500.7      \\
MERWIN     & 1926     &  1463.0      \\
Stone Soup & 2050     &  1450.6      \\
DecStar    & 1962     &  1437.3      \\
    \bottomrule
    \end{tabular}
\caption{Scores for state-of-the-art planners.}
\label{tab:baselines}
\end{center}
\end{table}

\begin{figure}[t]
\centering
\includegraphics[scale=.625]{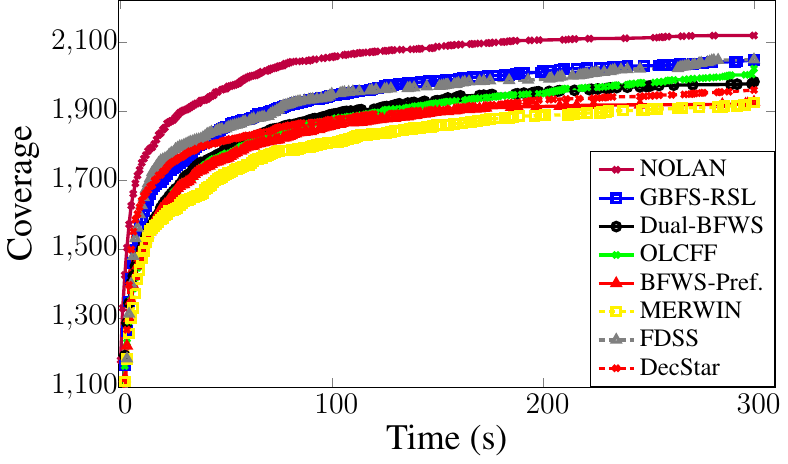}
\caption{Coverage over time for each state-of-the-art planner and NOLAN.}
\label{fig:sota}
\end{figure}

\section*{Consolidating the LAMA Open-Lists}

We started from LAMA-W($f_6$) and boiled the novelty-based open list down to
obtain LAMA-W($f_2^{\hlm}$), which then outperformed both LAMA and BFWS($f_6$).
But who says that all original parts of LAMA are beneficial in a planner with a
novelty-based open list? To evaluate this, we now consolidate the remaining
parts of LAMA-W($f_2^{\hlm}$) by removing subsets of LAMA open-lists from
LAMA-W($f_2^{\hlm}$).

As explained above, the original LAMA planner uses four open-lists: $[\hff,
\hff_+, \hlm, \hlm_+]$, while LAMA-W($f_2^{\hlm}$) has five: $[\hff, \hff_+,
\hlm, \hlm_+, f_2^{\hlm}]$. Table~\ref{tab:ablation-A} shows the effect of
removing combinations of of the original $4$ queues in LAMA from
LAMA-W($f_2^{\hlm}$).
The two strongest versions use three open-lists from LAMA plus the $f_2^{\hlm}$
open-list. In comparison to LAMA-W($f_2^{\hlm}$), removing the $\hff$ open-list
yields the highest coverage, while removing the $\hlm$ open-list improves
expansions and agile scores. Interestingly, the open-lists containing only
preferred operators ($\hff_+$ and $\hlm_+$) seem to be the most essential ones
in the planner.

For the rest of the paper, we use the version from Table~\ref{tab:ablation-B}
that has the highest agile and expansions scores (third column from the right).
It alternates between $[\hff, \hff_+, \hlm_+, f_2^{\hlm}]$. We call it NOLAN
because it uses (among other features) \acronym{NO}velty, \acronym{L}andmarks
and \acronym{A}lternatio\acronym{N}. The accumulated coverage over time by NOLAN
is almost identical LAMA-W($f_2^{\hlm}$) in Figure~\ref{fig:cactus-baseline}.

To better compare NOLAN to LAMA, we now analyze their results per domain.
Table~\ref{tab:domains} shows the coverage for the $26$ domains where the two
planners solve a different number of tasks. In $23$ out of the $26$, NOLAN
solves more tasks than LAMA. In more than half of these $23$ domains, NOLAN
solves two tasks or more compared to LAMA. On the flip side, LAMA has a higher
coverage in only $3$ domains. Two of them contain VisitAll tasks, where an agent
has to visit all cells in a rectangular grid. For these tasks, NOLAN frequently
runs out of memory. This is because VisitAll tasks have a huge number facts and
different $\hlm$ values, and the planner maintains a bit vector storing the seen
fact tuples for each $\hlm$ value. BFWS-based planners address this problem with
a simple strategy: if the estimated memory usage for storing the seen fact tuples
is larger than 2~GiB, they fall back to using $w = 1$. We leave addressing the
memory bottleneck for such tasks as future work.

\section*{Related Work}

Before we compare NOLAN to state-of-the-art agile planners, we discuss related
work, some of which forms the basis for some of the planners we compare to.

Balancing exploration and exploitation is a longstanding challenge in classical
planning
\cite{hoffmann-nebel-jair2001,richter-westphal-ipc2008,nakhost-mueller-ijcai2009,vidal-ipc2011,katz-et-al-icaps2017,asai-fukunaga-icaps2017,fickert-socs2018}.
In recent years, BFWS has emerged as the most successful approach for this
problem
\cite{lipovetzky-geffner-ecai2012,lipovetzky-geffner-aaai2017,frances-et-al-ijcai2017}.
We evaluate two representatives, BFWS-Preference and Dual-BFWS
\cite{frances-et-al-ipc2018}, in the next section.

Follow-up work combined BFWS with other search techniques. For example,
\citet{katz-et-al-icaps2017} combine the concept of novelty with heuristic
estimates, extending the definition of novelty by
\citet{shleyfman-et-al-ijcai2016} to take into account the heuristic value of
the states.
This allows them to quantify how novel a state is, so the search can be guided
directly by this value. However, as this metric is not goal-aware,
\citeauthor{katz-et-al-icaps2017} need to break ties using traditional
goal-aware functions. The MERWIN planner \cite{katz-et-al-ipc2018b}, that we
evaluate below, is based on these ideas.

Another successful approach is due to \citet{fickert-icaps2020}, who uses an
orthogonal approach to the one by \citeauthor{katz-et-al-icaps2017}: instead of using
novelty as the main guidance for the search, \citeauthor{fickert-icaps2020} uses
traditional heuristics to guide a greedy best-first search, and uses a lookahead
strategy to find states with lower heuristic values quickly. This lookahead
strategy is designed to reach states satisfying relaxed subgoals
\cite{lipovetzky-geffner-ecai2014}. To make the procedure efficient, he uses
novelty pruning \cite{lipovetzky-geffner-ecai2012,fickert-socs2018} to reduce
the number of evaluated states. \citeauthor{fickert-icaps2020} shows that there
is a synergy between the novelty-based lookahead and the $h^{C\textup{FF}}$
  heuristic \cite{fickert-hoffmann-icaps2017}, as the result from the lookahead
  can be used to trigger the refinement procedure of $h^{C\textup{FF}}$. This
    idea was also used in the OLCFF planner \cite{fickert-hoffmann-ipc2018} from
    the IPC 2018. We include OLCFF and an improved version of it, GBFS-RSL,
    in our empirical comparison below.

A simpler (but still efficient) idea was introduced by
\citet{correa-seipp-icaps2022} in the context of lifted planning. Besides
showing how to compute novelty over a lifted representation (where reachable
facts are not known in advance), they also introduce a planner that alternates
between a novelty-based open list and an open list ordered by traditional
heuristic estimates. While their planner has an overhead in most IPC domains due
to the first-order representation used, it is competitive with other planners for
larger tasks.

While many successful idea in classical planning follow the ``\emph{the more,
the merrier}'' idea \cite{roeger-helmert-icaps2010}, some lines of research go
in the opposite direction. For example, \citet{tuisov-katz-ijcai2021} define
\emph{novel} and \emph{non-novel operators}, and show that pruning non-novel
operators in preferred queues can increase coverage.

\section*{State-of-the-Art Agile Planners}

We now compare NOLAN to state-of-the-art agile planners which scored highly in
previous IPCs: BFWS-Preference \cite{frances-et-al-ipc2018}, the winner of the
Agile Track of IPC 2018; Dual-BFWS \cite{frances-et-al-ipc2018}, another
BFWS-based participant of IPC 2018; MERWIN \cite{katz-et-al-ipc2018b}, a planner
that combines the state-of-the-art satisficing planner Mercury
\cite{katz-hoffmann-ipc2014} with novelty heuristics
\cite{katz-et-al-icaps2017}; OLCFF \cite{fickert-hoffmann-ipc2018}, which
combines novelty pruning with the $h^\text{CFF}$ heuristic
\cite{hoffmann-fickert-icaps2015}; DecStar \cite{gnad-et-al-ipc2023b}, the
winner of the Agile Track at the IPC 2023; and Fast Downward Stone Soup 2023
\cite{buechner-et-al-ipc2023b}, the runner-up in the IPC 2023 Agile Track.
Furthermore, we also include GBFS-RSL \cite{fickert-icaps2020}, which is an
improved version of the OLCFF planner.

Table~\ref{tab:baselines} compares NOLAN with all state-of-the-art
planners.\footnote{We do not compare expansion scores for this experiment
because some planners perform multiple searches and some use look-aheads, which
skews the total number of expansions.} NOLAN has the highest coverage and agile
score. The relative difference in agile score to the second best planner,
GBFS-RSL, is about $13\%$, while the difference to the previous winners of the
IPC Agile Tracks is about $15\%$. As mentioned above, no other planner has an
agile score close to LAMA, even though some planners, e.g., Fast Downward Stone Soup
2023, use LAMA as one of their components.

Figure~\ref{fig:sota} shows the coverage over time for all of these planners.
The gap between NOLAN and all other methods is clearly visible. After $100$
seconds, NOLAN already solves more tasks than any other method at any point in
time.

\section*{Conclusions}

We showed how to combine two successful agile planners, LAMA and BFWS($f_6$).
The combination of both did not meet with success, and it yielded a worse
planner than any of its ingredients.  But by carefully removing features from
LAMA and the BFWS open-list, we obtained state-of-the-art performance. The
resulting planner NOLAN has a higher coverage and agile score than all other
evaluated planners.  These results raise the question of what happens if we
combine NOLAN with other exploration techniques \egcite{xie-et-al-aaai2014a}.

\section*{Acknowledgments}

Augusto B.\ Corrêa was funded by the Swiss National Science Foundation (SNSF) as
part of the project ``Lifted and Generalized Representations for Classical
Planning'' (LGR-Plan).
Jendrik Seipp was supported by the Wallenberg AI, Autonomous Systems and
Software Program (WASP) funded by the Knut and Alice Wallenberg Foundation.
Moreover, this work was partially supported by TAILOR, a project funded by EU
Horizon 2020 research and innovation programme under grant agreement
no.\ 952215.

\bibliography{abbrv,literatur,additional-bib,crossref}

\begin{thebibliography}{42}
\providecommand{\natexlab}[1]{#1}

\bibitem[{{Alc\'azar} and Torralba(2015)}]{alcazar-torralba-icaps2015}
{Alc\'azar}, V.; and Torralba, {\'A}. 2015.
\newblock A Reminder about the Importance of Computing and Exploiting
  Invariants in Planning.
\newblock In Brafman, R.; Domshlak, C.; Haslum, P.; and Zilberstein, S., eds.,
  \emph{Proceedings of the Twenty-Fifth International Conference on Automated
  Planning and Scheduling (ICAPS 2015)}, 2--6. AAAI Press.

\bibitem[{Asai and Fukunaga(2017{\natexlab{a}})}]{asai-fukunaga-icaps2017}
Asai, M.; and Fukunaga, A. 2017{\natexlab{a}}.
\newblock Exploration Among and Within Plateaus in Greedy Best-First Search.
\newblock In Barbulescu, L.; Frank, J.; Mausam; and Smith, S.~F., eds.,
  \emph{Proceedings of the Twenty-Seventh International Conference on Automated
  Planning and Scheduling (ICAPS 2017)}, 11--19. AAAI Press.

\bibitem[{Asai and Fukunaga(2017{\natexlab{b}})}]{asai-fukunaga-jair2017}
Asai, M.; and Fukunaga, A. 2017{\natexlab{b}}.
\newblock Tie-Breaking Strategies for Cost-Optimal Best First Search.
\newblock \emph{Journal of Artificial Intelligence Research}, 58: 67--121.

\bibitem[{Bonet and Geffner(2001)}]{bonet-geffner-aij2001}
Bonet, B.; and Geffner, H. 2001.
\newblock Planning as Heuristic Search.
\newblock \emph{Artificial Intelligence}, 129(1): 5--33.

\bibitem[{B{\"u}chner et~al.(2023)B{\"u}chner, Christen, Corr\^{e}a, Eriksson,
  Ferber, Seipp, and Sievers}]{buechner-et-al-ipc2023b}
B{\"u}chner, C.; Christen, R.; Corr\^{e}a, A.~B.; Eriksson, S.; Ferber, P.;
  Seipp, J.; and Sievers, S. 2023.
\newblock {Fast Downward Stone Soup} 2023.
\newblock In \emph{Tenth {I}nternational {P}lanning {C}ompetition ({IPC}-10):
  Planner Abstracts}.

\bibitem[{Bylander(1994)}]{bylander-aij1994}
Bylander, T. 1994.
\newblock The Computational Complexity of Propositional {STRIPS} Planning.
\newblock \emph{Artificial Intelligence}, 69(1--2): 165--204.

\bibitem[{Cenamor, {de la Rosa}, and
  Fern\'{a}ndez(2016)}]{cenamor-et-al-jair2016}
Cenamor, I.; {de la Rosa}, T.; and Fern\'{a}ndez, F. 2016.
\newblock The {IBaCoP} Planning System: Instance-Based Configured Portfolios.
\newblock \emph{Journal of Artificial Intelligence Research}, 56: 657--691.

\bibitem[{Corr\^{e}a et~al.(2023)Corr\^{e}a, Franc\`{e}s, Hecher, Longo, and
  Seipp}]{correa-et-al-ipc2023c}
Corr\^{e}a, A.~B.; Franc\`{e}s, G.; Hecher, M.; Longo, D.~M.; and Seipp, J.
  2023.
\newblock {Scorpion Maidu}: Width Search in the {Scorpion} Planning System.
\newblock In \emph{Tenth {I}nternational {P}lanning {C}ompetition ({IPC}-10):
  Planner Abstracts}.

\bibitem[{Corr{\^e}a and Seipp(2022)}]{correa-seipp-icaps2022}
Corr{\^e}a, A.~B.; and Seipp, J. 2022.
\newblock Best-First Width Search for Lifted Classical Planning.
\newblock In Thi{\'e}baux, S.; and Yeoh, W., eds., \emph{Proceedings of the
  Thirty-Second International Conference on Automated Planning and Scheduling
  (ICAPS 2022)}, 11--15. AAAI Press.

\bibitem[{Doran and Michie(1966)}]{doran-michie-rsl1966}
Doran, J.~E.; and Michie, D. 1966.
\newblock Experiments with the Graph Traverser program.
\newblock \emph{Proceedings of the Royal Society A}, 294: 235--259.

\bibitem[{Fickert(2018)}]{fickert-socs2018}
Fickert, M. 2018.
\newblock Making Hill-Climbing Great Again through Online Relaxation Refinement
  and Novelty Pruning.
\newblock In Bulitko, V.; and Storandt, S., eds., \emph{Proceedings of the 11th
  Annual Symposium on Combinatorial Search (SoCS 2018)}, 158--162. AAAI Press.

\bibitem[{Fickert(2020)}]{fickert-icaps2020}
Fickert, M. 2020.
\newblock A Novel Lookahead Strategy for Delete Relaxation Heuristics in Greedy
  Best-First Search.
\newblock In Beck, J.~C.; Karpas, E.; and Sohrabi, S., eds., \emph{Proceedings
  of the Thirtieth International Conference on Automated Planning and
  Scheduling (ICAPS 2020)}, 119--123. AAAI Press.

\bibitem[{Fickert and Hoffmann(2017)}]{fickert-hoffmann-icaps2017}
Fickert, M.; and Hoffmann, J. 2017.
\newblock Complete Local Search: Boosting Hill-Climbing through Online
  Relaxation Refinement.
\newblock In Barbulescu, L.; Frank, J.; Mausam; and Smith, S.~F., eds.,
  \emph{Proceedings of the Twenty-Seventh International Conference on Automated
  Planning and Scheduling (ICAPS 2017)}, 107--115. AAAI Press.

\bibitem[{Fickert and Hoffmann(2018)}]{fickert-hoffmann-ipc2018}
Fickert, M.; and Hoffmann, J. 2018.
\newblock {OLCFF}: Online-Learning {$h^\textup{CFF}$}.
\newblock In \emph{Ninth {I}nternational {P}lanning {C}ompetition ({IPC}-9):
  Planner Abstracts}, 17--19.

\bibitem[{Franc{\`e}s et~al.(2018)Franc{\`e}s, Geffner, Lipovetzky, and
  Ramir{\'e}z}]{frances-et-al-ipc2018}
Franc{\`e}s, G.; Geffner, H.; Lipovetzky, N.; and Ramir{\'e}z, M. 2018.
\newblock Best-First Width Search in the {IPC} 2018: Complete, Simulated, and
  Polynomial Variants.
\newblock In \emph{Ninth {I}nternational {P}lanning {C}ompetition ({IPC}-9):
  Planner Abstracts}, 23--27.

\bibitem[{Franc{\`e}s et~al.(2017)Franc{\`e}s, Ram{\'i}rez, Lipovetzky, and
  Geffner}]{frances-et-al-ijcai2017}
Franc{\`e}s, G.; Ram{\'i}rez, M.; Lipovetzky, N.; and Geffner, H. 2017.
\newblock Purely Declarative Action Representations are Overrated: Classical
  Planning with Simulators.
\newblock In Sierra, C., ed., \emph{Proceedings of the 26th International Joint
  Conference on Artificial Intelligence (IJCAI 2017)}, 4294--4301. IJCAI.

\bibitem[{Franco et~al.(2017)Franco, Torralba, Lelis, and
  Barley}]{franco-et-al-ijcai2017}
Franco, S.; Torralba, {\'A}.; Lelis, L. H.~S.; and Barley, M. 2017.
\newblock On Creating Complementary Pattern Databases.
\newblock In Sierra, C., ed., \emph{Proceedings of the 26th International Joint
  Conference on Artificial Intelligence (IJCAI 2017)}, 4302--4309. IJCAI.

\bibitem[{Gnad, Torralba, and Shleyfman(2023)}]{gnad-et-al-ipc2023b}
Gnad, D.; Torralba, {\'A}.; and Shleyfman, A. 2023.
\newblock {DecStar}-2023.
\newblock In \emph{Tenth {I}nternational {P}lanning {C}ompetition ({IPC}-10):
  Planner Abstracts}.

\bibitem[{Helmert(2004)}]{helmert-icaps2004}
Helmert, M. 2004.
\newblock A Planning Heuristic Based on Causal Graph Analysis.
\newblock In Zilberstein, S.; Koehler, J.; and Koenig, S., eds.,
  \emph{Proceedings of the Fourteenth International Conference on Automated
  Planning and Scheduling ({ICAPS} 2004)}, 161--170. AAAI Press.

\bibitem[{Helmert(2006)}]{helmert-jair2006}
Helmert, M. 2006.
\newblock The {Fast} {Downward} Planning System.
\newblock \emph{Journal of Artificial Intelligence Research}, 26: 191--246.

\bibitem[{Helmert, R{\"o}ger, and Karpas(2011)}]{helmert-et-al-icaps2011wspal}
Helmert, M.; R{\"o}ger, G.; and Karpas, E. 2011.
\newblock {Fast} {Downward} {Stone} {Soup}: A Baseline for Building Planner
  Portfolios.
\newblock In \emph{{ICAPS} 2011 Workshop on Planning and Learning}, 28--35.

\bibitem[{Hoffmann and Fickert(2015)}]{hoffmann-fickert-icaps2015}
Hoffmann, J.; and Fickert, M. 2015.
\newblock Explicit Conjunctions w/o Compilation: Computing
  {$h^\textup{FF}(\Pi^C)$} in Polynomial Time.
\newblock In Brafman, R.; Domshlak, C.; Haslum, P.; and Zilberstein, S., eds.,
  \emph{Proceedings of the Twenty-Fifth International Conference on Automated
  Planning and Scheduling (ICAPS 2015)}, 115--119. AAAI Press.

\bibitem[{Hoffmann and Nebel(2001)}]{hoffmann-nebel-jair2001}
Hoffmann, J.; and Nebel, B. 2001.
\newblock The {FF} Planning System: {Fast} Plan Generation Through Heuristic
  Search.
\newblock \emph{Journal of Artificial Intelligence Research}, 14: 253--302.

\bibitem[{Katz and Hoffmann(2014)}]{katz-hoffmann-ipc2014}
Katz, M.; and Hoffmann, J. 2014.
\newblock Mercury Planner: Pushing the Limits of Partial Delete Relaxation.
\newblock In \emph{Eighth {I}nternational {P}lanning {C}ompetition ({IPC}-8):
  Planner Abstracts}, 43--47.

\bibitem[{Katz et~al.(2017)Katz, Lipovetzky, Moshkovich, and
  Tuisov}]{katz-et-al-icaps2017}
Katz, M.; Lipovetzky, N.; Moshkovich, D.; and Tuisov, A. 2017.
\newblock Adapting Novelty to Classical Planning as Heuristic Search.
\newblock In Barbulescu, L.; Frank, J.; Mausam; and Smith, S.~F., eds.,
  \emph{Proceedings of the Twenty-Seventh International Conference on Automated
  Planning and Scheduling (ICAPS 2017)}, 172--180. AAAI Press.

\bibitem[{Katz et~al.(2018)Katz, Lipovetzky, Moshkovich, and
  Tuisov}]{katz-et-al-ipc2018b}
Katz, M.; Lipovetzky, N.; Moshkovich, D.; and Tuisov, A. 2018.
\newblock {MERWIN} Planner: Mercury Enchanced With Novelty Heuristic.
\newblock In \emph{Ninth {I}nternational {P}lanning {C}ompetition ({IPC}-9):
  Planner Abstracts}, 53--56.

\bibitem[{Lipovetzky and Geffner(2012)}]{lipovetzky-geffner-ecai2012}
Lipovetzky, N.; and Geffner, H. 2012.
\newblock Width and Serialization of Classical Planning Problems.
\newblock In {De Raedt}, L.; Bessiere, C.; Dubois, D.; Doherty, P.; Frasconi,
  P.; Heintz, F.; and Lucas, P., eds., \emph{Proceedings of the 20th {European}
  Conference on {Artificial} {Intelligence} ({ECAI} 2012)}, 540--545. IOS
  Press.

\bibitem[{Lipovetzky and Geffner(2014)}]{lipovetzky-geffner-ecai2014}
Lipovetzky, N.; and Geffner, H. 2014.
\newblock Width-based Algorithms for Classical Planning: New Results.
\newblock In Schaub, T.; Friedrich, G.; and O'Sullivan, B., eds.,
  \emph{Proceedings of the 21st {European} Conference on {Artificial}
  {Intelligence} ({ECAI} 2014)}, 1059--1060. IOS Press.

\bibitem[{Lipovetzky and Geffner(2017)}]{lipovetzky-geffner-aaai2017}
Lipovetzky, N.; and Geffner, H. 2017.
\newblock Best-First Width Search: Exploration and Exploitation in Classical
  Planning.
\newblock In Singh, S.; and Markovitch, S., eds., \emph{Proceedings of the
  Thirty-First {AAAI} Conference on Artificial Intelligence ({AAAI} 2017)},
  3590--3596. {AAAI} Press.

\bibitem[{Nakhost and M{\"u}ller(2009)}]{nakhost-mueller-ijcai2009}
Nakhost, H.; and M{\"u}ller, M. 2009.
\newblock {Monte}-{Carlo} Exploration for Deterministic Planning.
\newblock In Boutilier, C., ed., \emph{Proceedings of the 21st International
  Joint Conference on Artificial Intelligence ({IJCAI} 2009)}, 1766--1771. AAAI
  Press.

\bibitem[{Richter and Helmert(2009)}]{richter-helmert-icaps2009}
Richter, S.; and Helmert, M. 2009.
\newblock Preferred Operators and Deferred Evaluation in Satisficing Planning.
\newblock In Gerevini, A.; Howe, A.; Cesta, A.; and Refanidis, I., eds.,
  \emph{Proceedings of the Nineteenth International Conference on Automated
  Planning and Scheduling (ICAPS 2009)}, 273--280. AAAI Press.

\bibitem[{Richter, Helmert, and Westphal(2008)}]{richter-et-al-aaai2008}
Richter, S.; Helmert, M.; and Westphal, M. 2008.
\newblock Landmarks Revisited.
\newblock In \emph{Proceedings of the Twenty-Third {AAAI} Conference on
  Artificial Intelligence ({AAAI} 2008)}, 975--982. {AAAI} Press.

\bibitem[{Richter and Westphal(2008)}]{richter-westphal-ipc2008}
Richter, S.; and Westphal, M. 2008.
\newblock The {LAMA} Planner --- {Using} Landmark Counting in Heuristic Search.
\newblock IPC 2008 short papers,
  \url{http://ipc.informatik.uni-freiburg.de/Planners}.

\bibitem[{Richter and Westphal(2010)}]{richter-westphal-jair2010}
Richter, S.; and Westphal, M. 2010.
\newblock The {LAMA} Planner: Guiding Cost-Based Anytime Planning with
  Landmarks.
\newblock \emph{Journal of Artificial Intelligence Research}, 39: 127--177.

\bibitem[{Richter, Westphal, and Helmert(2011)}]{richter-et-al-ipc2011}
Richter, S.; Westphal, M.; and Helmert, M. 2011.
\newblock {LAMA} 2008 and 2011 (planner abstract).
\newblock In \emph{IPC 2011 Planner Abstracts}, 50--54.

\bibitem[{R{\"o}ger and Helmert(2010)}]{roeger-helmert-icaps2010}
R{\"o}ger, G.; and Helmert, M. 2010.
\newblock The More, the Merrier: Combining Heuristic Estimators for Satisficing
  Planning.
\newblock In Brafman, R.; Geffner, H.; Hoffmann, J.; and Kautz, H., eds.,
  \emph{Proceedings of the Twentieth International Conference on Automated
  Planning and Scheduling (ICAPS 2010)}, 246--249. AAAI Press.

\bibitem[{Seipp, Keller, and Helmert(2020)}]{seipp-et-al-jair2020}
Seipp, J.; Keller, T.; and Helmert, M. 2020.
\newblock Saturated Cost Partitioning for Optimal Classical Planning.
\newblock \emph{Journal of Artificial Intelligence Research}, 67: 129--167.

\bibitem[{Seipp et~al.(2017)Seipp, Pommerening, Sievers, and
  Helmert}]{seipp-et-al-zenodo2017}
Seipp, J.; Pommerening, F.; Sievers, S.; and Helmert, M. 2017.
\newblock {Downward} {Lab}.
\newblock \url{https://doi.org/10.5281/zenodo.790461}.

\bibitem[{Shleyfman, Tuisov, and Domshlak(2016)}]{shleyfman-et-al-ijcai2016}
Shleyfman, A.; Tuisov, A.; and Domshlak, C. 2016.
\newblock Blind Search for {Atari}-Like Online Planning Revisited.
\newblock In Kambhampati, S., ed., \emph{Proceedings of the 25th International
  Joint Conference on Artificial Intelligence (IJCAI 2016)}, 3251--3257. AAAI
  Press.

\bibitem[{Tuisov and Katz(2021)}]{tuisov-katz-ijcai2021}
Tuisov, A.; and Katz, M. 2021.
\newblock The Fewer the Merrier: Pruning Preferred Operators with Novelty.
\newblock In Zhou, Z.-H., ed., \emph{Proceedings of the 30th International
  Joint Conference on Artificial Intelligence (IJCAI 2021)}, 4190--4196. IJCAI.

\bibitem[{Vidal(2011)}]{vidal-ipc2011}
Vidal, V. 2011.
\newblock {YAHSP}2: Keep It Simple, Stupid.
\newblock In \emph{IPC 2011 Planner Abstracts}, 83--90.

\bibitem[{Xie et~al.(2014)Xie, M{\"u}ller, Holte, and
  Imai}]{xie-et-al-aaai2014a}
Xie, F.; M{\"u}ller, M.; Holte, R.~C.; and Imai, T. 2014.
\newblock Type-based Exploration with Multiple Search Queues for Satisficing
  Planning.
\newblock In Brodley, C.~E.; and Stone, P., eds., \emph{Proceedings of the
  Twenty-Eighth {AAAI} Conference on Artificial Intelligence ({AAAI} 2014)},
  2395--2401. {AAAI} Press.

\end{thebibliography}

\end{document}